\documentclass[journal]{IEEEtran}
\ifCLASSINFOpdf
\else
\fi
\usepackage[ruled,linesnumbered]{algorithm2e}
\usepackage{enumerate}

\usepackage{graphicx} 
\usepackage{amssymb}
\usepackage{epstopdf}
\usepackage{booktabs}
\usepackage[fleqn]{amsmath}
\usepackage{float}
\usepackage{setspace}
\usepackage{subfigure}
\usepackage{url}
\usepackage{color}
\usepackage{threeparttable} 
\usepackage[numbers,sort&compress]{natbib}

\begin{document}
%
\title{Random vector functional link neural network based ensemble deep
	learning for short-term load forecasting}


\author{\IEEEauthorblockN{Ruobin Gao\IEEEauthorrefmark{1},~\IEEEmembership{Member,~IEEE},
Liang Du\IEEEauthorrefmark{1},
P.N. Suganthan\IEEEauthorrefmark{2},~\IEEEmembership{Fellow,~IEEE},
Qin Zhou\IEEEauthorrefmark{1},
Kum Fai Yuen\IEEEauthorrefmark{1}}

\IEEEauthorblockA{\IEEEauthorrefmark{1}School of Civil and Environmental Engineering,
Nanyang Technological University, Singapore, Singapore}
\IEEEauthorblockA{\IEEEauthorrefmark{2}School of Electrical and Electronic Engineering,
	Nanyang Technological University, Singapore, Singapore}
\thanks{
Corresponding author: Ruobin Gao (email: gaor0009@e.ntu.edu.sg).}}


%
%

\markboth{}%
{Shell \MakeLowercase{\textit{et al.}}: Bare Demo of IEEEtran.cls for IEEE Journals}
%



\maketitle

\begin{abstract}
Electricity load forecasting is crucial for the power systems' planning and maintenance. However, its un-stationary and non-linear characteristics impose significant difficulties in anticipating future demand. This paper proposes a novel ensemble deep Random Vector Functional Link (edRVFL) network for electricity load forecasting. The weights of hidden layers are randomly initialized and kept fixed during the training process. The hidden layers are stacked to enforce deep representation learning. Then, the model generates the forecasts by ensembling the outputs of each layer. Moreover, we also propose to augment the random enhancement features by empirical wavelet transformation (EWT). The raw load data is decomposed by EWT in a walk-forward fashion, not introducing future data leakage problems in the decomposition process. Finally, all the sub-series generated by the EWT, including raw data, are fed into the edRVFL for forecasting purposes. The proposed model is evaluated on twenty publicly available time series from the Australian Energy Market Operator of the year 2020. The simulation results demonstrate the proposed model's superior performance over eleven forecasting methods in three error metrics and statistical tests on electricity load forecasting tasks. 

\end{abstract}


%
\IEEEpeerreviewmaketitle

\section{Introduction}
%
%
%
%
\IEEEPARstart{F}{orecasting} electricity load accurately benefits electric power system planning for maintenance and construction. After collecting raw electricity demand, a reliable forecasting model established on raw historical data can approximate how much electricity is expected in the future. Therefore, accurate forecasts help the supplier to decrease energy generation and expenses and plan the resources efficiently \cite{heydari2020short}. Furthermore, short-term load forecasting models assist electricity organizations in making opportune decisions in a data-driven fashion. As a result, developing novel and accurate forecasting models for short-term load is beneficial.
 
The electricity load forecasting is one kind of time series forecasting tasks. Anticipating the future using intelligent forecasting models is a well-developed field, where the models established from the historical data are used to extrapolate future values \cite{makridakis2008forecasting}. There are plentiful forecasting models, such as Auto-regressive integrated moving average (ARIMA) \cite{contreras2003arima}, fuzzy time series \cite{gao2020parsimonious}, support vector regression (SVR) \cite{chen2004load}, randomized neural networks \cite{ren2016random}, hybrid models \cite{gao2020robust,gao2021walk,qiu2017empirical,ren2014comparative}, ensemble learning \cite{qiu2017oblique,qiu2018ensemble} and deep learning models \cite{almalaq2017review}. Accurate and reliable forecasts of electricity load is a challenging and significant problem for the electric power domain. In the field of load forecasting domain, the methods can be classified into three categories (i) statistical models, (ii) computational intelligence models and (ii) hybrid models. The statistical models, such as ARIMA \cite{contreras2003arima} and exponential smoothing \cite{taylor2011short}, are computationally efficient and theoretically solid, but their performance is not outstanding. The second huge branch is the computational intelligence models including fuzzy system \cite{ali2020load,gao2020robust}, SVR \cite{chen2004load}, shallow artificial neural networks (ANN) \cite{ren2016random} and deep learning \cite{almalaq2017review,shi2017deep,hafeez2020electric,fekri2021deep,chitalia2020robust,qiu2014ensemble}. In \cite{shi2017deep}, a pooling deep recurrent neural network (RNN) is proposed to overcome the over-fitting problem caused by deep structures. A deep factored conditional restricted Boltzmann machine (FCRBM) whose parameters are optimized via a genetic wind-driven optimization (GWDO) for load forecasting is proposed in \cite{hafeez2020electric}. In \cite{fekri2021deep}, online tuning is utilized to update the deep RNN when the performance degrades. Several deep RNNs are evaluated for load forecasting in \cite{chitalia2020robust}, where the input is selected from various weather and scheduled related variables. The last category, hybrid models, includes the combination of feature extraction blocks and several forecasting models to form a single model. For example, the empirical mode decomposition (EMD) is utilized to extract modes from the load and then deep belief network (DBN) is implemented to forecast each mode in \cite{qiu2017empirical}. Empirical wavelet transformation (EWT) is applied to decompose the load data into sub-series in a walk-forward fashion and then the concatenation of raw data and sub-series are fed into a random vector functional link (RVFL) network for forecasting purposes \cite{gao2021walk}. 

Neural networks are popular models for load forecasting due to their high accuracy and strong ability to handle non-linearity. The deep learning models \cite{almalaq2017review,shi2017deep,hafeez2020electric,fekri2021deep,chitalia2020robust,qiu2014ensemble} succeed in forecasting short-term load accurately because of their hierarchical structures which learn a meaningful representation of the input data. However, most fully trained deep learning models suffer from huge computation burdens. Therefore, this paper proposes a fast ensemble deep learning algorithm for short-term load forecasting. The proposed model inherits the advantages of ensemble learning and deep learning without imposing much computational burden at the same time. This paper investigates the forecasting ability of a special kind of randomized deep neural networks, the deep RVFL network, whose training is fast. Ensemble learning techniques are combined with the deep RVFL to reduce the uncertainty caused by a single model. Since the deep RVFL's hidden features are randomly generated and remained fixed during the training process, the EWT is utilized to extract features with different frequencies to augment the deep RVFL's random features. Recently, the universal function approximation ability of the RVFL network is proved in \cite{needell2020random}. This paper uses EWT to decompose the raw data in a walk-forward fashion which is different from decomposing the whole time series altogether \cite{qiu2017empirical,ren2014comparative,ren2014novel,qiu2018ensemble}. The future data are not involved in the walk-forward decomposition process. Therefore there is no data leakage problem in terms of forecasting. 

The novel characteristics of the proposed model are summarized as follows: 
\begin{enumerate}[1.]
	\item This paper implements the edRVFL for short-term load forecasting for the first time. The mean and median computations are used as ensemble approaches which are different from the edRVFL for classification \cite{shi2021random}. 
\item The EWT is combined with the edRVFL as a feature engineering block to augment the random features. Furthermore, the EWT is conducted in a walk-forward fashion to avoid future data leakage problems. Finally, two novel hybrid forecasting models based on walk-forward EWT and edRVFL are proposed for short-term load forecasting.
\item The hyper-parameters of the proposed model are optimized in a layer-wise fashion. The succeeding layers are based on the optimized previous layer's features. Therefore, each layer has its suitable hyper-parameters and does not degrade the performance.
\item The proposed model is compared with various benchmark models from statistical ones to state-of-the-art models on twenty load time series. Three error metrics and two statistical tests are conducted for precise comparisons. The statistical tests demonstrate the proposed model's superiority both in a group-wise and pair-wise fashion.
	
\end{enumerate}

The remainder of this paper is organized as follows: Section \uppercase\expandafter{\romannumeral2} describes the methodologies and the proposed model in detail. We first describe the EWT and the walk-forward decomposition. Then, the ensemble deep RVFL and its combination with the walk-forward EWT is presented. Section \uppercase\expandafter{\romannumeral3} presents the experimental step-up and the results. Finally, conclusions are drawn and potential future directions are discussed in Section \uppercase\expandafter{\romannumeral4}.
 


\section{Methodology}
This section describes the methodologies in detail. First, we introduce the EWT and the walk-forward decomposition procedure. Then, we describe the ensemble deep RVFL network and the proposed model.
\subsection{Empirical wavelet transformation}
 The EWT is an automatic signal decomposition algorithm with solid theoretical foundations and remarkable effectiveness in decomposing non-stationary time series data \cite{gilles2013empirical}. Unlike discrete wavelet transform (DWT) and EMD \citep{flandrin2004empirical}, EWT precisely investigates the time series in the Fourier domain after fast Fourier transform (FFT). It realizes the spectrum separation using band-pass filtering with the data-driven filter banks. 

\begin{figure}[htbp]
	\centering
	\includegraphics[width=.5\textwidth]{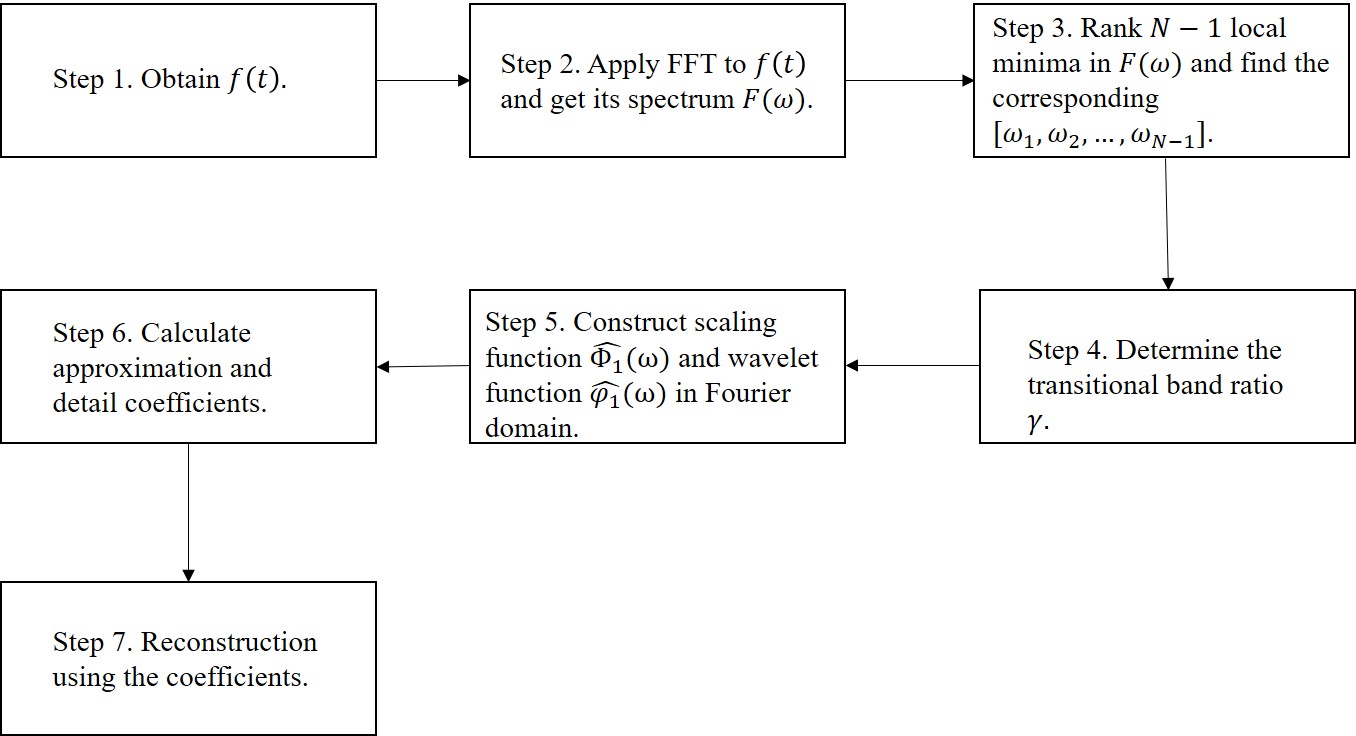}
	\caption{EWT implementation.}
	\label{fig:EWT workflow}
\end{figure}

Figure \ref{fig:EWT workflow} shows the EWT's regular procedures. In the EWT, limited freedom is provided for selecting wavelets. The algorithm employs Littlewood-Paley and Meyer's wavelets because of the analytic accessibility of the Fourier domain's closed-form expression  \citep{spencer1994ten}. In \cite{gilles2013empirical}, the formulations of these band-pass filters are denoted by Equations \ref{eq:EWT_1} and \ref{eq:EWT_2}
\begin{equation}
	\centering
	\label{eq:EWT_1}
	\hat{\phi}_{n}(\omega) = 
	\begin{cases}
		1& \makebox[1pt][r]{\text{if $\left|\omega\right| \leq (1-\gamma)\omega_{n}$}}\\
		
		\cos\left[\frac{\pi}{2}\beta\left(\frac{1}{2\gamma\omega_{n}}\left(\left|\omega\right|- (1-\gamma)\omega_{n}|\right)\right)\right]
		& \\
		\quad&\makebox[60pt][r]{\text{if $(1-\gamma)\omega_{n} \leq \left|\omega \right| \leq (1+\gamma)\omega_{n}$ }}\\
		
		0  & \text{otherwise,}\\
	\end{cases}
\end{equation} 

\begin{equation}
	\centering
	\label{eq:EWT_2}
	\hat{\psi}_{n}(\omega) = 
	\begin{cases}
		
		1& \makebox[12pt][r]{\text{if $(1+\gamma)\omega_{n} \leq \left|\omega \right| \leq (1-\gamma)\omega_{n+1}$ }}\\
		
		\cos\left[\frac{\pi}{2}\zeta\left(\frac{1}{2\gamma\omega_{n+1}}\left(\left|\omega\right|- (1-\gamma)\omega_{n+1}|\right)\right)\right]
		& \\
	\quad	&\makebox[22pt][r]{\text{if $(1-\gamma)\omega_{n+1} \leq \left|\omega \right| \leq (1+\gamma)\omega_{n+1}$ }}\\
		
		\sin\left[\frac{\pi}{2}\zeta\left(\frac{1}{2\gamma\omega_{n}}\left(\left|\omega\right|- (1-\gamma)\omega_{n}|\right)\right)\right]
		& \\
		\quad&\makebox[0.001pt][r]{\text{if $(1-\gamma)\omega_{n} \leq \left|\omega \right| \leq (1+\gamma)\omega_{n}$}}\\
		
		0  & \makebox[2pt][r]{\text{otherwise,}}\\
		
	\end{cases}
\end{equation}
with a transitional band width parameter $\gamma$ satisfying $\gamma \le \min_{n} \frac{\omega_{n+1}-\omega_{n}}{\omega_{n+1}+\omega_{n}}$. The most common function $\zeta(x)$ in Equation \ref{eq:EWT_1} and \ref{eq:EWT_2} is presented in Equation \ref{eq:Beta}. This empowers the formulated empirical scaling and wavelet function
$\{\hat{\phi}_{1}(\omega),\{\hat{\psi}_{n}(\omega)\}_{n=1}^{N} \}$ to be a tight frame of $L^{2}(\mathbb{R})$ \citep{casazza2000art}.

\begin{equation}
	\label{eq:Beta}
	\centering
	\beta(x)=x^{4}(35-84x+70x^2-20x^3)
\end{equation}
It can be observed that $\{\hat{\phi}_{1}(\omega),\{\hat{\psi}_{n}(\omega)\}_{n=1}^{N} \}$ are used as band-pass filters centered at assorted center frequencies. 
\subsection{Walk-forward decomposition}
Plentiful works utilize signal decomposition techniques as a feature engineering block for the forecasting algorithms \cite{qiu2017empirical,ren2014comparative,ghelardoni2013energy,gao2020robust,yang2018time,gao2021walk,huang2021new}, however, most do not implement the decomposition in a proper way \cite{gao2021walk,huang2021new}. As mentioned in \cite{gao2020robust,gao2021walk,huang2021new}, direct application of signal decomposition algorithm to the whole time series causes the data leakage problem in terms of forecasting. The decomposed data are actually the output from convolution operations and the future data definitely are involved during the convolution. Therefore, decomposition of the whole time series is incorrect and improper, especially for establishing forecasting models.

Some solutions are proposed to avoid the future data leakage problem for decomposition-based forecasting models, such as the data-driven padding \cite{gao2020robust}, moving window strategy \cite{huang2021new} and walk-forward decomposition \cite{gao2021walk}. The data-driven padding approach is to train a simple learning algorithm which aims at padding its forecast to the end of the time series \cite{gao2020robust}. The moving window strategy only decomposes the data located in the window (order) and then the decomposed series are fed into forecasting models \cite{huang2021new}. Different from the moving window strategy, only part of the decomposed sub-series are used as input in the walk-forward decomposition. The moving window strategy is a subset of the walk-forward decomposition. When the order is equal to the window, the moving window strategy and the walk-forward decomposition are the same. 

This paper adopts the walk-forward decomposition for the EWT. The walk-forward EWT decomposes the data in a rolling window $w$, which consists of $x(t-1),x(t-2),... ...x(t-w)$, into $k$ scales with the aim to predict $x(t)$. Then only the last order data points are used as input for the forecasting model. Therefore, only historical observations are involved both in the decomposition process and the model' training.
\subsection{Ensemble deep RVFL}
Inspired by the deep representation learning, the deep RVFL is an extension of the RVFL with a shallow structure \cite{shi2021random}. The deep RVFL is established by stacking multiple enhancement layers to achieve deep representation learning. The clean data are fed into each enhancement layer to guide the random features' generation. In this fashion, the enhancement features of hidden layers are generated based on the information from the clean data and the features from the previous layer. A diverse set of features is generated with the help of hierarchical structures. Ensemble learning is introduced into the deep RVFL architecture to formulate the ensemble deep RVFL (edRVFL). Different from the popular deep learning models with a single output layer, the edRVFL trains multiple output layers based on all the hidden features. Finally, the forecasts from all output layers are combined for forecasting.
 
\begin{figure*}[htbp]
	\centering
	\includegraphics[width=\textwidth]{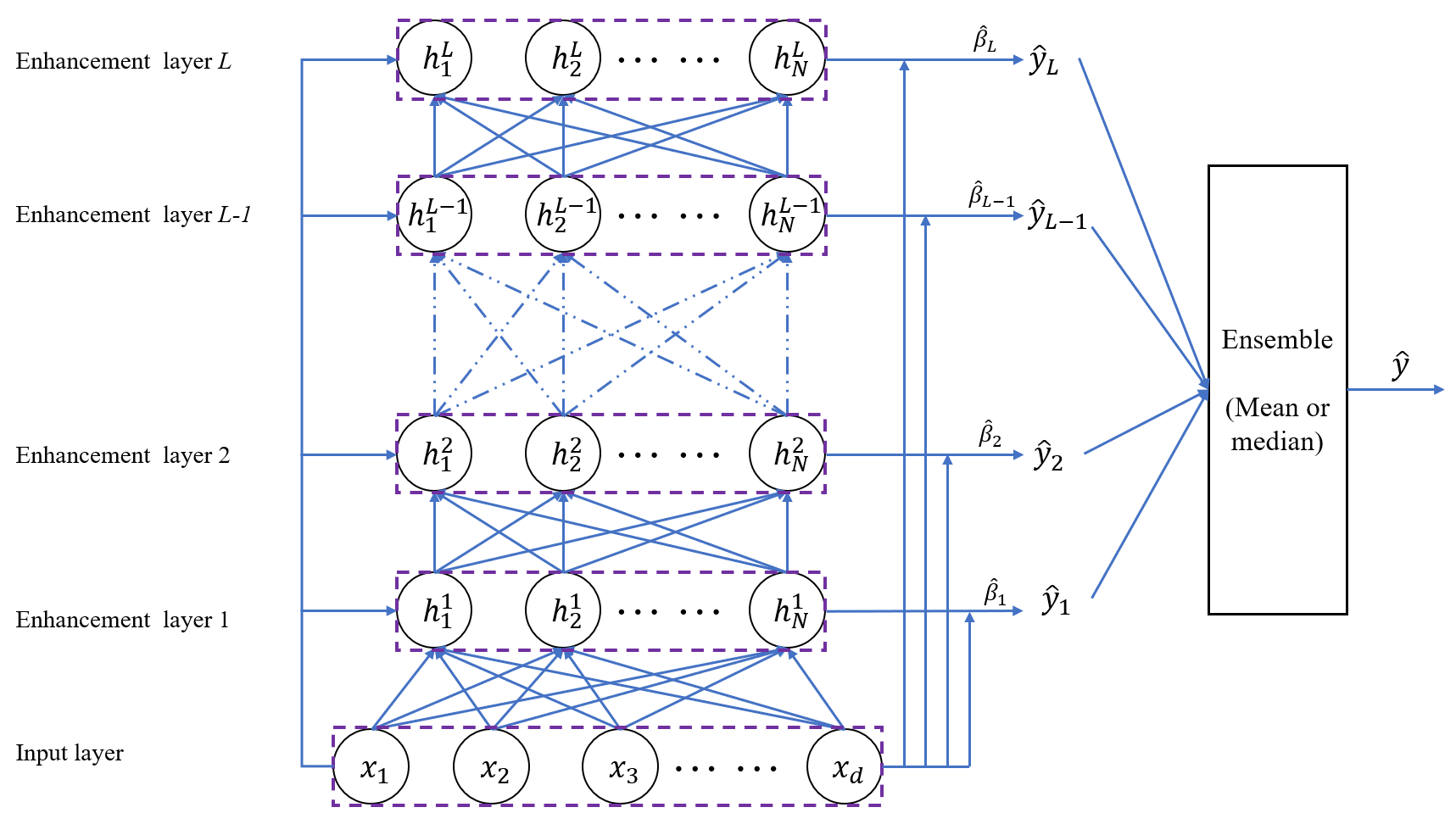}
	\caption{Architecture of the edRVFL.}
	\label{fig:Architecture of the edRVFL}
\end{figure*}

For the sake of presentation simplicity, we only present the edRVFL with a structure of $L$ enhancement layers and there are $N$ enhancement nodes in each layer. Figure \ref{fig:Architecture of the edRVFL} shows the architecture of the edRVFL network. 
Suppose that the input data is $\mathbf{X} \in \mathbb{R}^{ n \times d}$, where $n$ and $d$ represent the number of samples and feature dimension, respectively. $d$ is the time lag (order) for the time series forecasting model. The features generated by the first enhancement layer are defined as

\begin{equation}
	\mathbf{H^{1}}=g(\mathbf{XW_{1}}),
\end{equation}

where $\mathbf{W_{1}} \in \mathbb{R}^{d \times N}$ represents the weight vector of the first enhancement layer, $\mathbf{H^{1}} \in \mathbb{R}^{n \times N}$ denotes the enhancement features and $g()$ is a non-linear activation function. The readers can refer to \cite{zhang2016comprehensive} for a comprehensive evaluation of different activation functions.  
Then, for the deeper enhancement layer $l$, the enhancement features can be computed as

\begin{equation}
   \mathbf{H^{\textit{l}}}=g(\mathbf{[H^{\textit{l}-1},X]W_{\textit{l}}}),
\end{equation}

where $\mathbf{W_{\textit{l}}} \in \mathbb{R}^{(d+N)\times N}$ and $\mathbf{H^{\textit{l}}}\in \mathbb{R}^{n \times N} $. The enhancement weight vectors $\mathbf{W_{1}}$ and $\mathbf{W_{\textit{l}}}$ are randomly initialized and remained fixed during training. 

The edRVFL computes the output weights by splitting the task into $l$ small tasks. The output weights are calculated separately for each layer. There are several differences from using the last layer's features and all layers' features for decisions. Most deep learning models only use the last layer's features for decisions, however, the information from the intermediate features is lost. Using all layers' features requires a computation on the feature matrix with a huge dimension. Moreover, both of the above architectures only train one network, but our method benefits from the ensemble approach, which reduces the uncertainty of a single model. 

The loss function of $l^{th}$ enhancement layer is defined as
\begin{equation}
	Loss_{\textit{l}}=||\mathbf{[H^{\textit{l}},X]}\beta_{\textit{l}}-Y||^{2}+\lambda ||\beta_{\textit{l}}||^{2}, 
\end{equation}
	where $\beta_{\textit{l}}$ denotes the output vector of $l^{th}$ layer and $\lambda$ is the regularization parameter. The minimization of $Loss_{\textit{l}}$ can be solved via a closed-form solution based on ridge regression \cite{saunders1998ridge}. 
\begin{equation}
	\label{eq:ridge regression}
	\beta_{\textit{l}}=(\mathbf{D^{T}D}+\lambda \mathbf{I})^{-1}\mathbf{D}^{T}Y,
\end{equation}
where $\mathbf{D}=\mathbf{[H^{\textit{l}},X]}$. After computing all $\beta_{\textit{l}}$, the deep network can output $L$ forecasts. The final forecast is an ensemble of all outputs. Any forecast combination approach can be applied to this procedure \cite{timmermann2006forecast}. According to the suggestions in \cite{timmermann2006forecast}, the mean or median operation is always likely to improve the forecast combination's performance. Therefore, we use the mean and median as the combination operator. Correspondingly, two different edRVFLs are proposed, the Mea-edRVFL and Med-edRVFL.
\subsection{EWT-edRVFL}
The model EWT-edRVFL consists of two blocks, the walk-forward EWT decomposition and the edRVFL. The walk-forward EWT is first applied to the load data to extract some features in a causal fashion. Then the raw data concatenated with the sub-series are fed into the edRVFL with $L$ enhancement layers for learning purposes. The output weights $\beta_{\textit{l}}$ of the $l^{th}$ enhancement layer are computed according to Equation \ref{eq:ridge regression}. Finally, we ensemble the $L$ forecasts with mean or median operation to obtain the output $\hat{y}$. Correspondingly, two different EWT-edRVFLs are proposed, the EWTMea-edRVFL and EWTMed-edRVFL.

Since the higher enhancement layer's performance depends on the lower ones', the hyper-parameters of the whole model are tuned in a layer-wise fashion. Once the shallow layer's hyper-parameters are determined, then they are fixed and the cross-validation approach is applied to the next layer. Layer-wise cross-validation offers a different set of hyper-parameters for each layer. Therefore each enhancement layer has its own regularization parameter, which helps the overall edRVFL learns a diverse set of output layers.
\section{Empirical study}
This section presents the empirical study on twenty load time series collected from the Australian Energy Market Operator (AEMO). First, we briefly introduce the data' characteristics and pre-processing steps. Then, the benchmark models and hyper-parameter optimization are described. Finally, the simulation results are shown, and discussions are conducted.

\subsection{Data and its nature}
Table \ref{tab:Descriptive statistics.} summarizes the descriptive statistics of the twenty load time series. These load data are collected from the states of South Australia (SA), Queensland (QLD), New South Wales (NSW), Victoria (VIC), and Tasmania (TAS) of the year 2020, which is significantly affected by Covid-19. Four months, January, April, July, and October are selected to reflect the four seasons' characteristics as in \cite{gao2021walk,jalali2021novel,qiu2017empirical}. The data are recorded every half an hour. Therefore, there are 48 data points per day. 

A suitable and correct data pre-processing approach helps the machine learning model generate accurate outputs. We utilize the max-min normalization to pre-process the raw data. We assume that the maximum and minimum of the training set are $x_{max}$ and $x_{min}$, respectively. The data are transformed into the range [0,1] using the following equation:
\begin{equation}
	x_{normalized}=\frac{x-x_{min}}{x_{max}-x_{min}}
\end{equation}
where $x_{normalized}$ and $x$ represent the normalized and original time series, respectively.

All datasets are split into three sets, the training, validation and test set, to adopt the cross-validation \cite{bergmeir2012use}. The validation and test set account for 10\% and 20\% of the dataset, respectively. The remaining data are used as the training set. 
 \begin{table}[htbp]
 	\centering
 	\caption{Descriptive statistics.}
 	\label{tab:Descriptive statistics.}
 	\resizebox{0.5\textwidth}{!}{
 	\begin{tabular}{lllllllll}
 		\hline
 		Location& Month   & Max      & Min     & Median  & Mean    & Std     & Skewness & Kurtosis \\
 		\hline
 		SA  & Jan     & 3085.49  & 440.54  & 1212.79 & 1268.80 & 427.93  & 1.26     & 2.60     \\
 		& Apr   & 1841.85  & 503.67  & 1177.78 & 1161.61 & 248.31  & -0.33    & -0.37    \\
 		& Jul    & 2383.18  & 765.27  & 1489.76 & 1514.57 & 338.45  & 0.26     & -0.59    \\
 		& Oct & 1955.46  & 288.92  & 1140.50 & 1095.25 & 266.31  & -0.55    & 0.21     \\
 		QLD & Jan     & 9620.91  & 5407.70 & 6824.81 & 6941.23 & 949.16  & 0.44     & -0.65    \\
 		& Apr   & 7722.78  & 4480.52 & 5783.49 & 5916.37 & 693.05  & 0.60     & -0.48    \\
 		& Jul    & 8148.44  & 4216.62 & 5783.27 & 5925.44 & 812.46  & 0.35     & -0.87    \\
 		& Oct & 7646.61  & 3921.39 & 5503.29 & 5673.93 & 746.37  & 0.41     & -0.59    \\
 		NSW & Jan     & 13330.14 & 5765.85 & 8053.13 & 8264.22 & 1535.24 & 0.85     & 0.42     \\
 		& Apr   & 9471.04  & 5384.58 & 6983.91 & 6926.61 & 792.43  & 0.20     & -0.58    \\
 		& Jul    & 11739.02 & 5678.37 & 8670.19 & 8690.30 & 1247.70 & 0.17     & -0.75    \\
 		& Oct & 9324.77  & 5221.13 & 6999.92 & 6955.32 & 771.00  & 0.01     & -0.62    \\
 		VIC & Jan     & 9507.26  & 3060.58 & 4565.41 & 4765.55 & 1017.14 & 1.82     & 4.39     \\
 		& Apr   & 6515.96  & 3094.45 & 4453.18 & 4485.45 & 632.63  & 0.29     & -0.42    \\
 		& Jul    & 7354.11  & 3816.70 & 5497.73 & 5514.65 & 832.99  & 0.04     & -0.92    \\
 		& Oct & 6142.91  & 2975.43 & 4325.26 & 4379.82 & 587.84  & 0.27     & -0.53    \\
 		TAS & Jan     & 1298.63  & 794.25  & 1036.17 & 1040.35 & 84.44   & 0.09     & -0.26    \\
 		& Apr   & 1379.49  & 843.31  & 1087.11 & 1093.91 & 113.14  & 0.22     & -0.71    \\
 		& Jul    & 1597.64  & 887.09  & 1240.32 & 1246.55 & 151.24  & 0.08     & -0.86    \\
 		& Oct & 1447.61  & 842.78  & 1068.39 & 1087.26 & 112.91  & 0.47     & -0.33   \\
 		\hline
 	\end{tabular}}
 \end{table}

\subsection{Results and discussion}
Three forecasting error metrics are employed to appraise the accuracy of these models. The first error metric is the regular root mean square error (RMSE) whose definition is 
\begin{equation}
	RMSE=\sqrt{\frac{1}{L_{test}}\sum_{j=1}^{L_{test}}(\hat{x_{j}}-x_{j})^2},
\end{equation}
where $L_{test}$ is the size of the test set, $x_j$ and $\hat{x_j}$ are the raw data and predictions. The second error metric implemented in the paper is the mean absolute scaled error (MASE) \cite{hyndman2006another}. The definition of MASE is
\begin{equation}
	MASE=mean(\frac{\hat{x_{j}}-x_{j}}{\frac{1}{L_{train}-1}\sum_{t=2}^{L_{train}}|x_{t}-x_{t-1}|}),
\end{equation}
where $L_{train}$ represents the size of training set. The denominator of MASE is the mean absolute error of the in-sample naive forecast. The third error metric is the Mean Absolute Percentage Error (MAPE) whose definition is
\begin{equation}
	MAPE=\frac{1}{L_{test}}\sum_{j=1}^{L_{test}}|\frac{\hat{x_{j}}-x_{j}}{x_{j}}|.
\end{equation}

We compare the proposed model with many classical and state-of-the-art models. These models are Persistence model \cite{makridakis2008forecasting}, ARIMA \cite{contreras2003arima}, SVR \cite{chen2004load}, MLP \cite{almalaq2017review}, LSTM \cite{kong2017short}, Temporal CNN (TCN) \cite{baiempirical}, hybrid EWT fuzzy cognitive map (FCM) learned with SVR (EWTFCMSVR) \cite{gao2020robust}, Wavelet High-order FCM (WHFCM) \cite{yang2018time}, Laplacian ESN (LapESN) \cite{han2017laplacian}, EWTRVFL \cite{gao2021walk} and RVFL \cite{ren2016random}. The previous one day, 48 data points are used as input for all the models as in \cite{gao2021walk}. To achieve a fair comparison, all models' hyper-parameters are optimized by cross-validation. The hyper-parameter search space is presented in Table \ref{tab:Hyper-parameter search space for the benchmark models.}. The decomposition level for the walk-forward EWT is set to 2 according to the conclusion and suggestions in \cite{gao2021walk}. Some parameters are not involved in the optimization process and they are set to the same values for all the relevant models, which include the batch size equals to 32, learning rate equals to 0.001 and epochs equal to 200.

\begin{figure}[htbp]
	\centering
	
	\includegraphics[width=.45\textwidth]{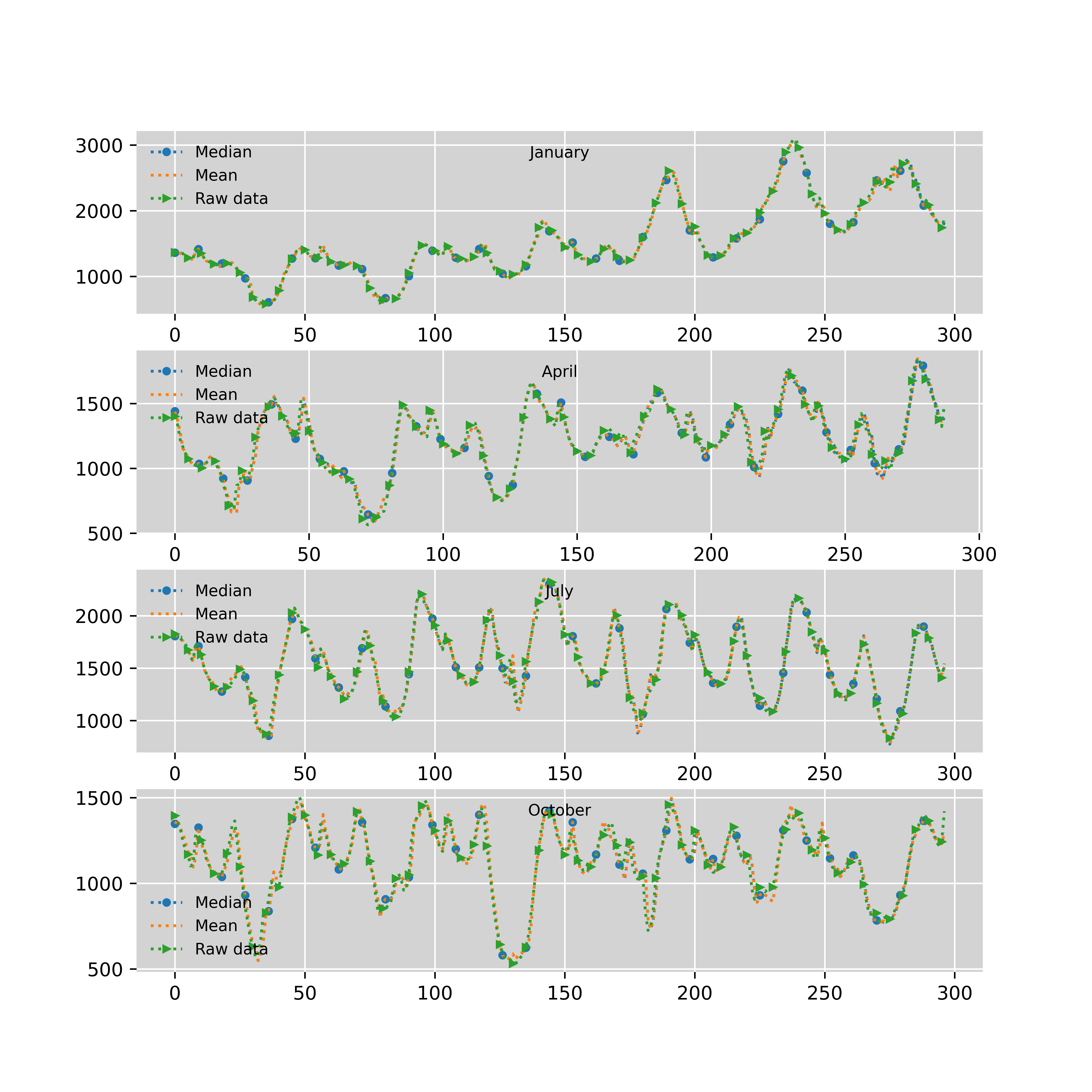}
	
	\caption{Comparisons of raw data and forecasts for the SA dataset.}
	\label{fig:test_curve_SA}
\end{figure}

\begin{figure}[htbp]
	\centering
	
	\includegraphics[width=.45\textwidth]{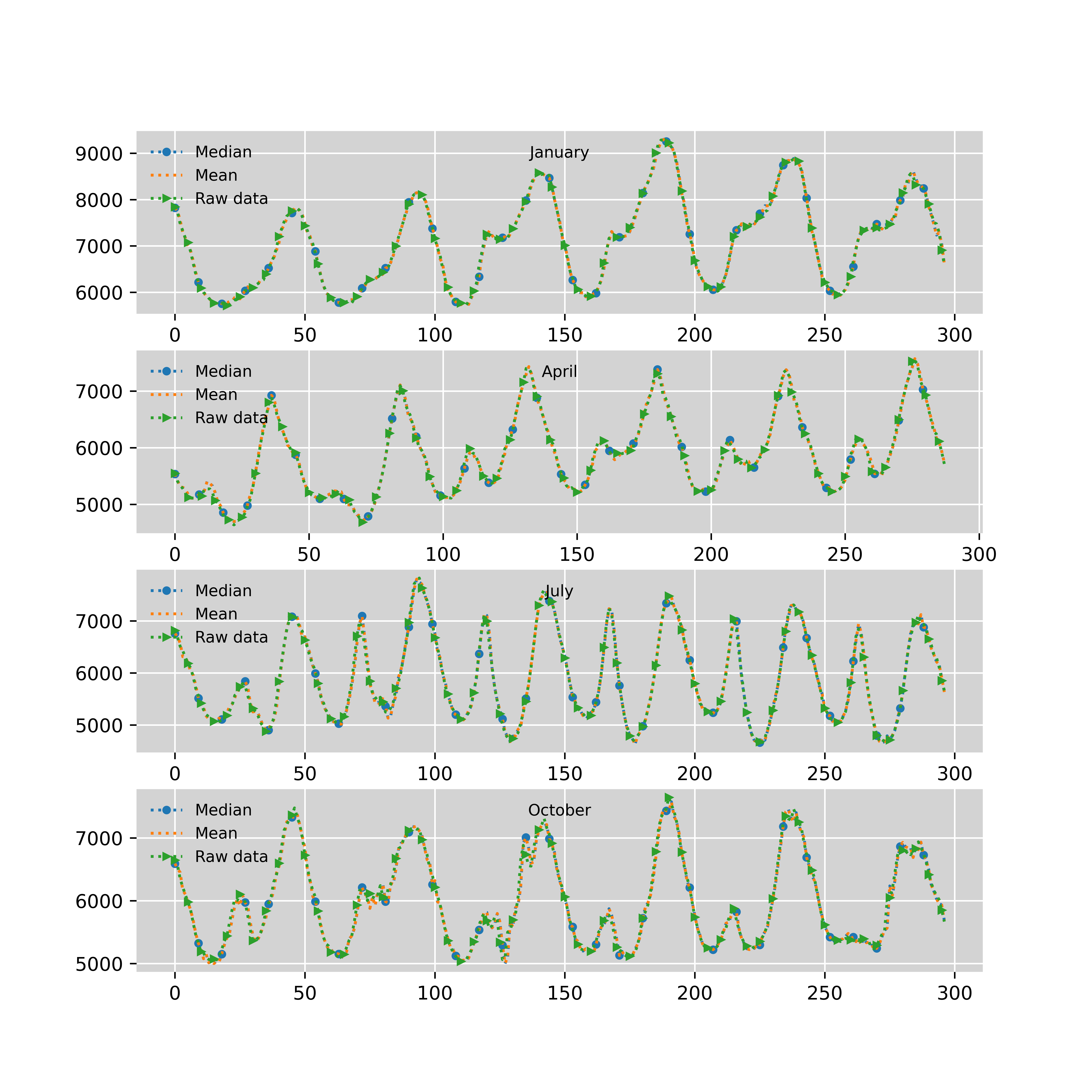}
	
	\caption{Comparisons of raw data and forecasts for the QLD dataset.}
	\label{fig:test_curve_QLD}
\end{figure}

\begin{figure}[htbp]
	\centering
	
	\includegraphics[width=.45\textwidth]{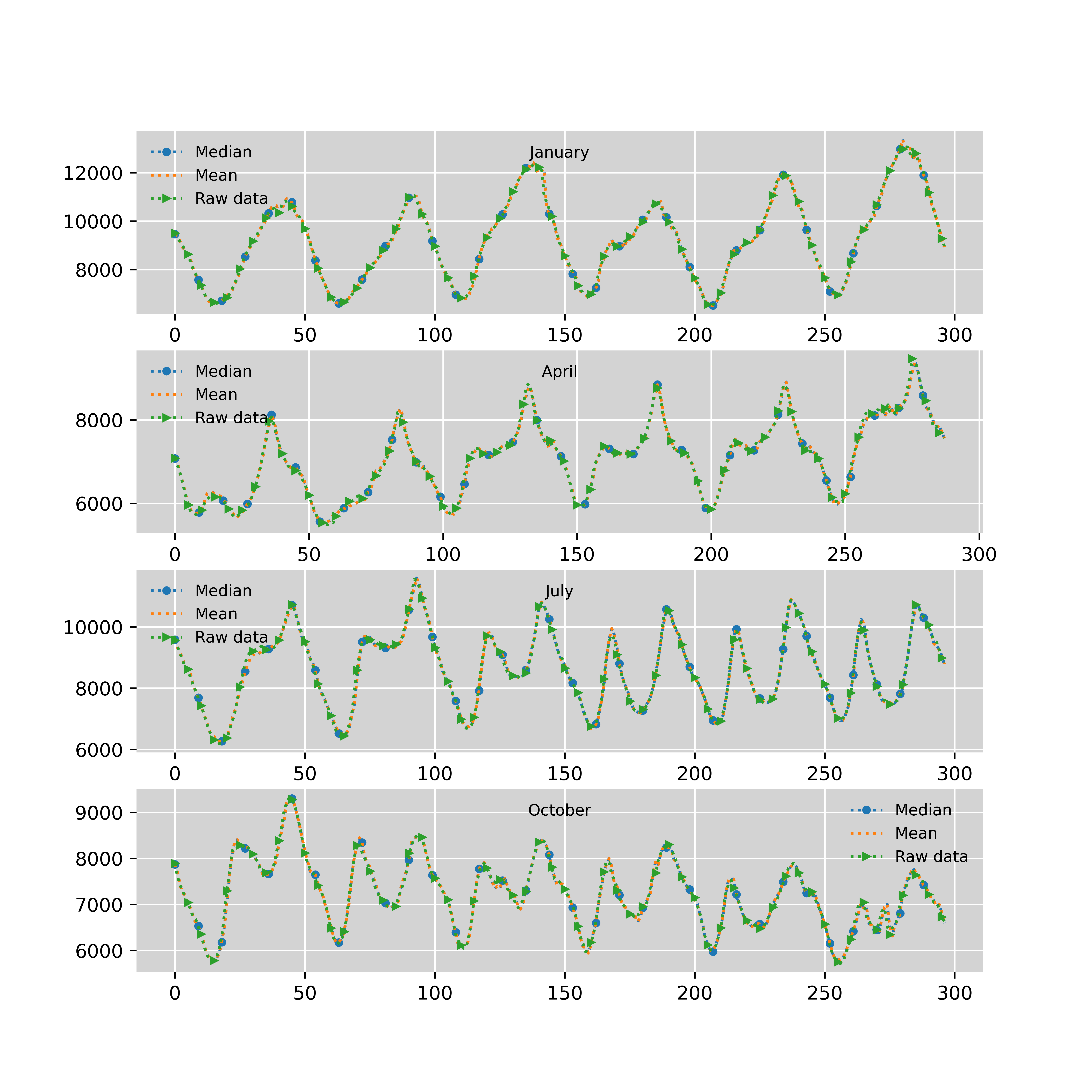}
	
	\caption{Comparisons of raw data and forecasts for the NSW dataset.}
	\label{fig:test_curve_NSW}
\end{figure}

\begin{figure}[htbp]
	\centering
	
	\includegraphics[width=.5\textwidth]{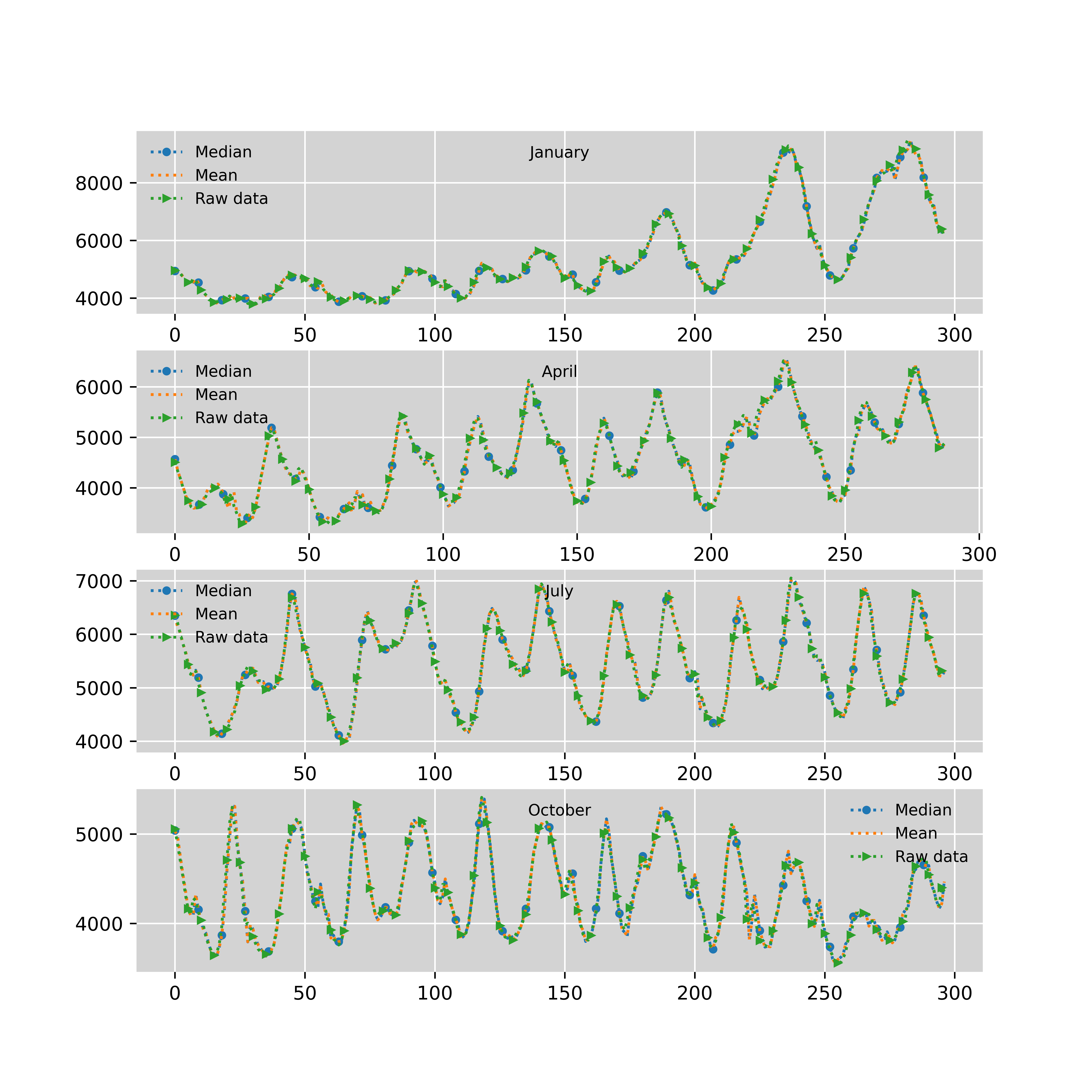}
	
	\caption{Comparisons of raw data and forecasts for the VIC dataset.}
	\label{fig:test_curve_VIC}
\end{figure}

\begin{figure}[htbp]
	\centering
	
	\includegraphics[width=.45\textwidth]{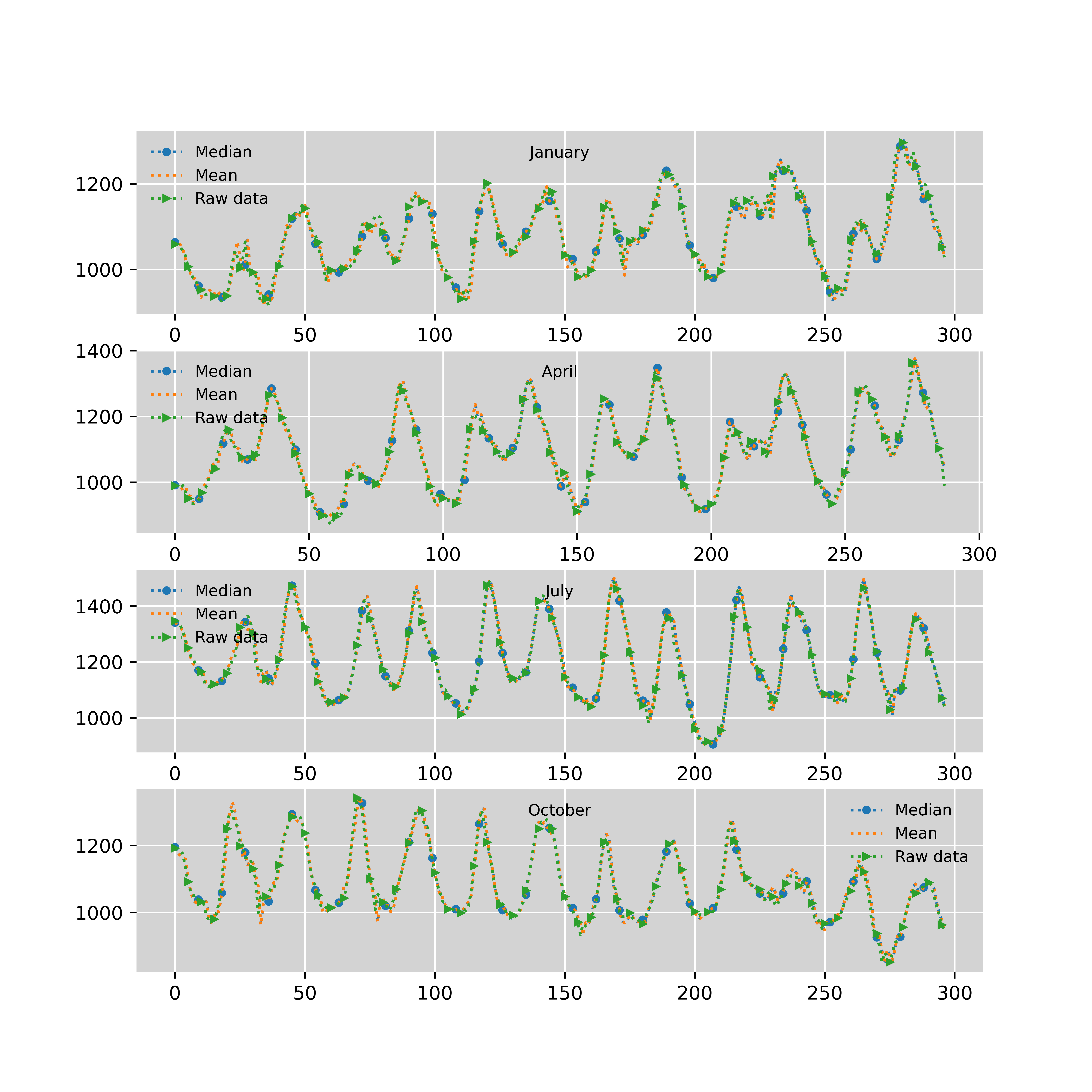}
	
	\caption{Comparisons of raw data and forecasts for the TAS dataset.}
	\label{fig:test_curve_TAS}
\end{figure}

\begin{figure}[htbp]
	\centering
	\subfigure[]{
		\includegraphics[width=.4\textwidth]{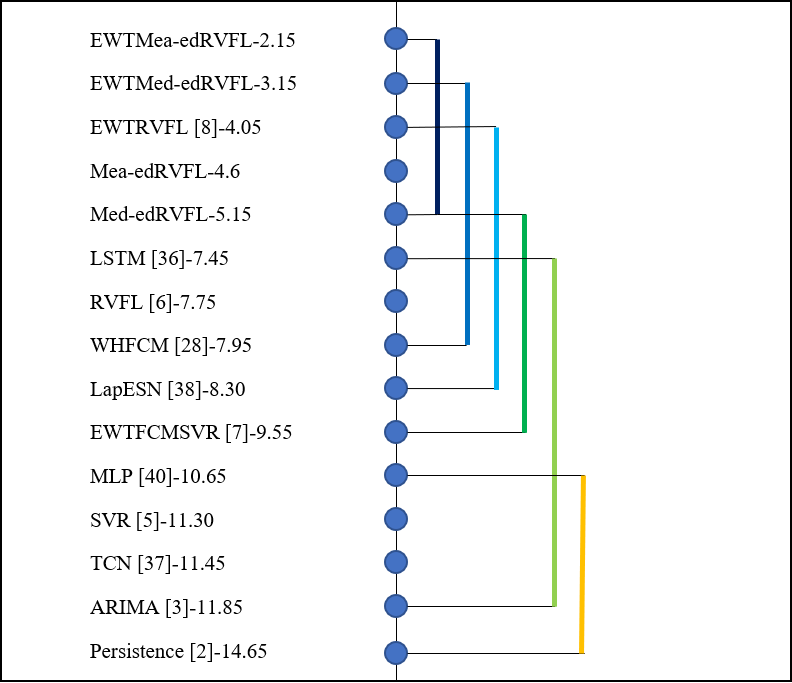}
	}
	\quad
	\subfigure[]{
		\includegraphics[width=.4\textwidth]{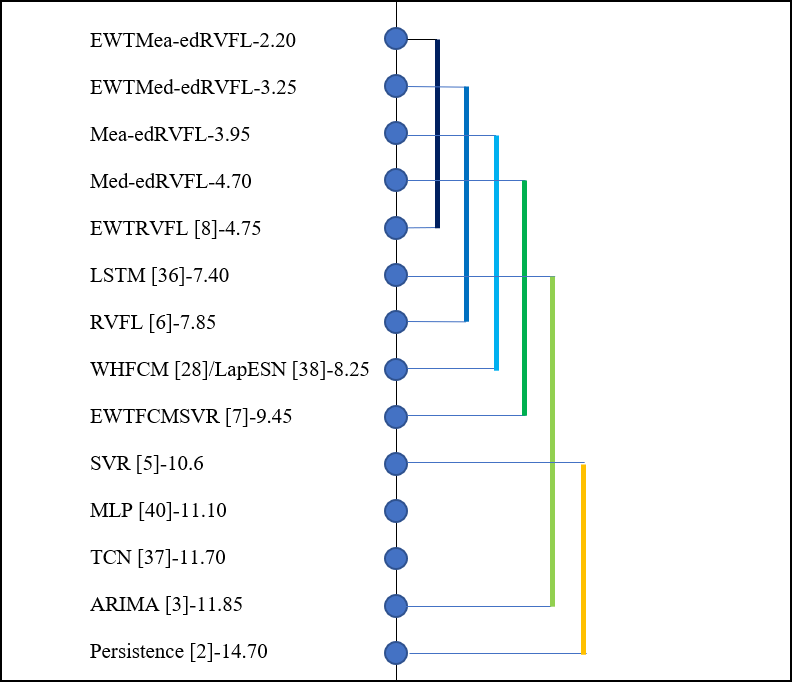}
	}
	\quad
	\subfigure[]{
		\includegraphics[width=.4\textwidth]{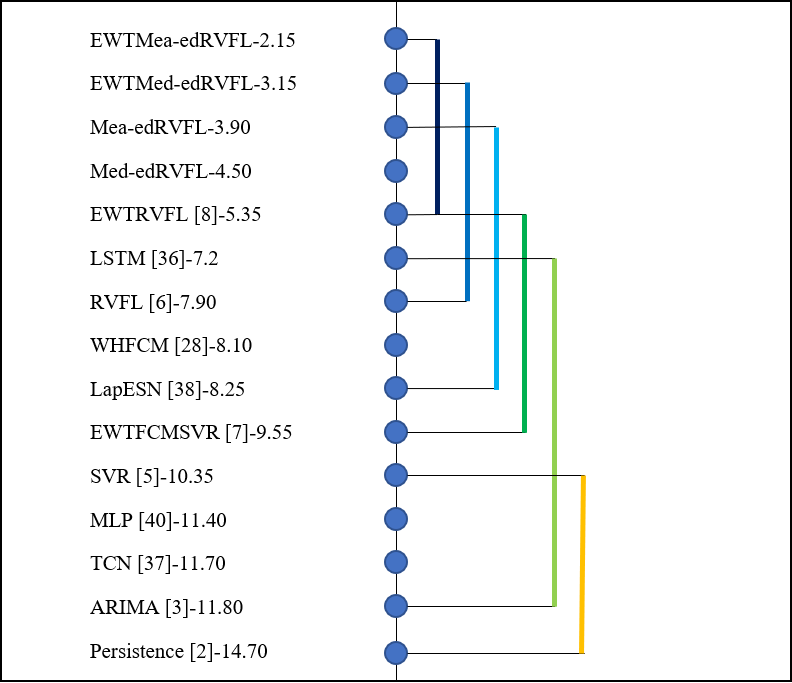}
	}
	\quad
	\caption{Nemenyi testing results for load forecasting based on: (a) RMSE, (b) MASE and (c) MAPE. The critical distance is 4.50. The Friedman p-values are: (a) 6.41e-45, (b) 3.02e-37 and (c) 8.48e-40. }
	\label{fig:Nem test}
\end{figure}

Tables \ref{tab:Comparative results in terms of RMSE.}, \ref{tab:Comparative results in terms of MASE.} and \ref{tab:Comparative results in terms of MAPE.} summarize the performance on the test sets. The numbers in bold represent the corresponding model's performance is the best on the specific time series. Figures \ref{fig:test_curve_SA}, \ref{fig:test_curve_QLD}, \ref{fig:test_curve_NSW}, \ref{fig:test_curve_VIC} and \ref{fig:test_curve_TAS} present the comparison of raw data and the forecasts generated by the proposed model. It is clear to find that the proposed model anticipates future trends, cycles, and fluctuations accurately. Statistical tests are implemented to investigate the difference among all the models further. We first implement the Friedman test, and the $p$-value is smaller than 0.05, which represents that these forecasting models are significantly different on these twenty datasets. Therefore, a post-hoc Nemenyi test is utilized to distinguish them \citep{demvsar2006statistical}. The critical distance of the Nemenyi test is calculated by:
\begin{equation}
	CD=q_{\alpha}\sqrt{\frac{k(k+1)}{6N_{d}}}
\end{equation}
where $q_{\alpha}$ is the critical value coming from the studentized range statistic divided by $\sqrt{2}$, $k$ represents the number of models and $N_{d}$ is the number of datasets \citep{demvsar2006statistical}. Figure \ref{fig:Nem test} represents the Nemenyi test results. The figures show that the models that achieve excellent performance are at the top, whereas the model with the worst performance is at the bottom. Some consistent conclusions can be drawn from the Nemenyi test results of three error metrics. The Persistence method is the tailender because it learns nothing about the patterns. ARIMA is a penultimate because of its simple linear structure. The LSTM model outperforms many benchmark models except the EWTRVFL and the model proposed in this paper. Figure \ref{fig:Nem test} demonstrates the superiority of the proposed models because they are always at the top. Another finding is that the edRVFL with mean ensemble operator is better than the median operator. A pair-wise Nemenyi post-hoc statistical comparison is further conducted and the $p$ values are shown in Tables \ref{tab:Pairwise RMSE.}, \ref{tab:Pairwise MASE.} and \ref{tab:Pairwise MAPE.}. The $p$ values smaller than 0.05 indicate that the two corresponding models are significantly different. The negative one in the diagonal positions represent that it is meaningless to compare the model with itself. The proposed EWTMea-edRVFL is significantly different from Persistence, ARIMA, SVR, MLP, LSTM, TCN, EWT-FCM-SVR, WHFCM, LapESN, and RVFL. The Mea-edRVFL and Med-edRVFL do not show significant superiority over LSTM, WHFCM, Lap ESN, RVFL, EWT-RVFL, and the EWT-based edRVFL models. 
\begin{table}[htbp]
		\centering
	\caption{Hyper-parameter search space for the benchmark models.}
	\label{tab:Hyper-parameter search space for the benchmark models.}
	\resizebox{.5\textwidth}{!}{
	\begin{tabular}{lll}
		\hline
		Model           & Parameter                & Values                                                    \\
		\hline
		ARIMA           & $p/q$                      & [1,2,3]                                                     \\
		
		& $d$                        & 0,1                                                       \\
		\hline
		SVR             & $C$                        & $[2^{-10},2^{0}]$ \\
		& $\epsilon$  & [0.001,0.01,0.1]                                            \\
		& 
		Radius                   & [0.001,0.01,0.1]                                            \\
		\hline
		MLP             & Hidden nodes             & {[}2,4,8,16,32{]}                                         \\
		& Layers                   & {[}1,2,3{]}                                               \\
		& Optimizer                & Adam                                                      \\
		& Activation               & $Relu$                                                      \\
		\hline
		LSTM            & Hidden nodes             & {[}2,4,8,16,32{]}                                         \\
		& Layers                   & {[}1,2,3{]}                                               \\
		& Optimizer                & Adam                                                      \\
		& Activation               & $Tanh$                                                      \\
		\hline
		TCN             & Filters                  & {[}2,4,8,16,32{]}                                         \\
		& Kernel size              & 2                                                         \\
		& Optimizer                & Adam                                                      \\
		& Activation               & $Relu$                                                      \\
		\hline
		EWTFCMSVR & Concepts                 & {[}2,6{]}                                                 \\
		WHFCM& Regularization           & $[2^{0},2^{-8}]$                                          \\
		\hline
		LapESN          & Reservoir size           & {[}50,200,50{]}                                           \\
		& Spectral radius          & {[}0.96,0.98{]}                                           \\
		
		& Input scalings           & {[}0.001,0.01,0.1{]}                                      \\
		\\
		\hline
		RVFL            & Ennhancement nodes       & {[}50,200,50{]}                                           \\
		Proposed        & Regularization parameter & $[2^{-4},2^{-8},0]$   \\
		\hline                                          
	\end{tabular}}
\end{table}
\begin{table*}[htbp]
	\centering
	\caption{Comparative results in terms of RMSE.}
	\label{tab:Comparative results in terms of RMSE.}
	\resizebox{\textwidth}{!}{
	\begin{tabular}{lllllllllllllllll}
		\hline
		Location &    & Persistence \cite{makridakis2008forecasting} & ARIMA \cite{contreras2003arima}    & SVR \cite{chen2004load}      & MLP \cite{kandil2006efficient}      & LSTM \cite{kong2017short}           & TCN \cite{baiempirical}      & EWTFCMSVR \cite{gao2020robust}        & WHFCM \cite{yang2018time}    & LapESN \cite{han2017laplacian}   & RVFL \cite{ren2016random}     & EWTRVFL \cite{gao2021walk}           & Med-edRVFL  & Mea-edRVFL  & EWTMed-edRVFL       & EWTMea-edRVFL       \\
		\hline
		SA  & Jan  & 72.8870     & 55.6380  & 66.9695  & 54.0868  & 59.6167          & 56.8684  & 70.2112          & 45.6521  & 52.2596  & 56.9607  & 45.0455           & 48.6848  & 48.7193  & \textbf{43.3276} & 43.6756          \\
		& Apr  & 67.9854     & 55.6386  & 52.7960  & 49.5188  & 55.7887          & 58.4276  & 111.5263         & 50.7253  & 53.0914  & 51.7494  & \textbf{47.3257}  & 50.1305  & 49.8541  & 48.2897          & 47.9759          \\
		& Jul  & 96.1804     & 61.5945  & 50.4778  & 56.4437  & 46.0799          & 58.9092  & \textbf{43.7050} & 45.9180  & 53.9056  & 50.9046  & 45.5713           & 49.1807  & 48.6929  & 45.0094          & 44.9430          \\
		& Oct  & 66.4943     & 51.9143  & 48.7796  & 54.9934  & 44.5717          & 50.2702  & 63.0079          & 45.1588  & 46.1959  & 45.3673  & \textbf{44.2800}  & 44.7918  & 44.6774  & 46.0017          & 45.7415          \\
		QLD & Jan  & 149.3016    & 72.1442  & 101.8223 & 84.6203  & 63.4356          & 69.3515  & 99.0970          & 58.6258  & 58.4367  & 56.2622  & \textbf{54.1518}  & 55.4811  & 55.4332  & 54.8145          & 54.5508          \\
		& Apr  & 135.1574    & 72.4100  & 57.1544  & 60.1787  & \textbf{49.3209} & 75.5726  & 149.5419         & 58.7142  & 60.0888  & 56.8695  & 53.2141           & 54.6669  & 54.1989  & 51.6416          & 51.0361          \\
		& Jul  & 217.7307    & 92.4658  & 73.7379  & 79.8500  & 63.6131          & 73.6105  & 64.9643          & 75.7170  & 69.3350  & 66.0357  & \textbf{62.2991}  & 64.3635  & 64.1440  & 63.3489          & 63.0091          \\
		& Oct  & 149.8869    & 101.2928 & 122.5457 & 106.1602 & 102.8157         & 102.7253 & 193.6453         & 92.7304  & 93.5727  & 93.7343  & 92.5209           & 91.4382  & 91.3391  & 91.4528          & \textbf{91.2787} \\
		NSW & Jan  & 241.6598    & 128.5224 & 176.6366 & 177.7787 & 124.1342         & 123.5894 & 124.7843         & 125.0457 & 121.5456 & 119.2369 & \textbf{118.7250} & 119.3599 & 119.1618 & 120.6610         & 120.3967         \\
		& Apr  & 170.5394    & 107.0291 & 123.2308 & 82.3968  & 109.8911         & 114.7059 & 262.1209         & 75.4579  & 86.3372  & 82.9668  & \textbf{68.5926}  & 81.1944  & 80.5445  & 74.4880          & 73.7032          \\
		& Jul & 294.9618    & 131.8007 & 130.3933 & 119.4671 & 100.2549         & 98.0032  & \textbf{85.5482} & 95.0363  & 103.8141 & 105.9918 & 120.4936          & 96.4300  & 99.9890  & 94.2159          & 93.9689          \\
		&Oct & 179.3761    & 97.9060  & 147.0332 & 96.8127  & 105.0375         & 125.7355 & 99.5157          & 85.9031  & 87.3090  & 87.9294  & 83.3434           & 83.5984  & 83.4924  & 81.4568          & \textbf{81.1915} \\
		VIC & Jan & 166.0274    & 107.4523 & 476.5141 & 105.5127 & 160.8402         & 167.1842 & \textbf{80.5299} & 96.0986  & 96.8277  & 99.3172  & 89.0404           & 96.9804  & 98.4364  & 93.2445          & 93.6646          \\
		& Apr & 161.6524    & 95.0628  & 112.7885 & 91.8794  & 90.1323          & 103.5377 & 157.6802         & 79.5648  & 79.3548  & 85.3221  & 85.0635           & 77.2368  & 77.3312  & 76.7201          & \textbf{76.4103} \\
		& Jul & 202.3882    & 100.1305 & 76.3694  & 73.9571  & 66.8791          & 86.8470  & 234.3873         & 77.2782  & 77.6613  & 71.9779  & 68.2234           & 68.4402  & 67.8317  & 66.9922          & \textbf{66.4860} \\
		& Oct & 146.7197    & 93.5497  & 85.8821  & 84.7936  & 78.0583          & 95.2157  & \textbf{68.0958} & 77.1013  & 75.3540  & 76.0608  & 70.7279           & 72.0342  & 71.8692  & 71.1336          & 70.5075          \\
		TAS & Jan & 22.6897     & 18.8835  & 21.7117  & 20.3779  & \textbf{17.7090} & 19.9012  & 18.6469          & 18.7898  & 18.3987  & 18.3759  & 18.3235           & 18.2444  & 18.2500  & 18.2270          & 18.2175          \\
		& Apr & 29.5110     & 20.4644  & 17.2503  & 25.9389  & 18.1222          & 19.1709  & 17.5378          & 20.0259  & 18.1358  & 17.5151  & 17.6300           & 17.2185  & 17.1762  & 17.0570          & \textbf{17.0362} \\
		& Jul & 41.6062     & 24.1888  & 22.5443  & 22.4608  & 20.4853          & 22.6558  & 20.1255          & 24.9275  & 20.9395  & 21.7029  & 20.8809           & 20.2539  & 20.1882  & 19.6646          & \textbf{19.5957} \\
		& Oct & 30.8810     & 21.3638  & 19.4400  & 20.1470  & 19.3222          & 21.0611  & 20.0869          & 20.8855  & 19.9530  & 19.6488  & 19.4225           & 19.4212  & 19.3918  & 19.2082          & \textbf{19.1757}
\\
\hline	
\end{tabular}
}
\end{table*}

\begin{table*}[htbp]
	\centering
	\caption{Comparative results in terms of MASE.}
	\label{tab:Comparative results in terms of MASE.}
	\resizebox{\textwidth}{!}{
	\begin{tabular}{lllllllllllllllll}
		\hline
		Location & Month   & Persistence \cite{makridakis2008forecasting}& ARIMA \cite{contreras2003arima} & SVR \cite{chen2004load}   & MLP \cite{kandil2006efficient}   & LSTM \cite{kong2017short}          & TCN \cite{baiempirical}    & EWTFCMSVR \cite{gao2020robust}       & WHFCM \cite{yang2018time} & LapESN \cite{han2017laplacian} & RVFL \cite{ren2016random}   & EWTRVFL \cite{gao2021walk}       & Med-edRVFL & Mea-edRVFL & EWTMed-edRVFL      & EWTMea-edRVFL      \\
		\hline
		SA  & Jan     & 1.2552      & 0.8463 & 0.9083 & 0.8405 & 0.8355          & 0.8471 & 1.0406          & 0.7138 & 0.7733 & 0.8153 & 0.6966          & 0.7188  & 0.7209  & \textbf{0.6720} & 0.6772          \\
		& Apr   & 1.1203      & 0.8195 & 0.7782 & 0.7581 & 0.8278          & 0.8801 & 1.7619          & 0.8120 & 0.7971 & 0.8048 & 0.7271          & 0.7513  & 0.7477  & 0.7328          & \textbf{0.7283} \\
		& Jul    & 1.1060      & 0.5701 & 0.4610 & 0.5598 & 0.4125          & 0.5624 & \textbf{0.4049} & 0.4437 & 0.5319 & 0.5103 & 0.4411          & 0.4705  & 0.4656  & 0.4322          & 0.4319          \\
		& Oct & 1.0056      & 0.7204 & 0.6215 & 0.8209 & 0.6088          & 0.6858 & 0.8455          & 0.6309 & 0.6299 & 0.6353 & \textbf{0.6303} & 0.6159  & 0.6148  & 0.6494          & 0.6454          \\
		QLD & Jan     & 1.0560      & 0.4847 & 0.6849 & 0.6120 & 0.4403          & 0.4635 & 0.4630          & 0.3981 & 0.4017 & 0.3898 & 0.3776          & 0.3838  & 0.3835  & 0.3736          & \textbf{0.3718} \\
		& Apr   & 1.0272      & 0.5268 & 0.3905 & 0.4328 & 0.3295          & 0.5529 & 1.1220          & 0.4356 & 0.4401 & 0.4121 & 0.3871          & 0.3892  & 0.3857  & 0.3745          & \textbf{0.3702} \\
		& Jul    & 1.1261      & 0.4237 & 0.3345 & 0.3696 & 0.3015          & 0.3432 & 0.3085          & 0.3544 & 0.3248 & 0.3146 & 0.2898          & 0.3060  & 0.3052  & 0.2977          & \textbf{0.2957} \\
		& Oct & 1.0274      & 0.6096 & 0.7358 & 0.6492 & 0.6073          & 0.6036 & 1.3169          & 0.5511 & 0.5521 & 0.5646 & 0.5576          & 0.5444  & 0.5431  & 0.5432          & \textbf{0.5422} \\
		NSW & Jan     & 1.4312      & 0.5671 & 0.8771 & 0.9933 & 0.5749          & 0.5624 & 0.5576          & 0.5445 & 0.5562 & 0.5439 & \textbf{0.5404} & 0.5363  & 0.5358  & 0.5426          & 0.5418          \\
		& Apr   & 1.0095      & 0.6026 & 0.5571 & 0.4514 & 0.5353          & 0.6150 & 1.5863          & 0.4308 & 0.4851 & 0.4540 & \textbf{0.3834} & 0.4393  & 0.4358  & 0.4101          & 0.4056          \\
		& Jul    & 0.9287      & 0.3917 & 0.3625 & 0.3415 & 0.3018          & 0.2910 & \textbf{0.2551} & 0.2842 & 0.3078 & 0.3074 & 0.3355          & 0.2750  & 0.2820  & 0.2693          & 0.2680          \\
		& Oct & 1.0979      & 0.5425 & 0.7185 & 0.5264 & 0.5644          & 0.6956 & 0.5590          & 0.4746 & 0.4696 & 0.4790 & 0.4571          & 0.4504  & 0.4497  & 0.4340          & \textbf{0.4326} \\
		VIC & Jan     & 1.3105      & 0.7993 & 2.3222 & 0.8405 & 1.0803          & 1.0792 & \textbf{0.6126} & 0.7456 & 0.7330 & 0.7268 & 0.6341          & 0.7153  & 0.7193  & 0.6875          & 0.6874          \\
		& Apr   & 1.1833      & 0.6260 & 0.7401 & 0.6363 & 0.5785          & 0.6515 & 0.9166          & 0.5284 & 0.5260 & 0.5611 & 0.5613          & 0.5078  & 0.5080  & 0.5014          & \textbf{0.4994} \\
		& Jul    & 1.0659      & 0.4864 & 0.3698 & 0.3608 & 0.3264          & 0.4103 & 1.2698          & 0.3729 & 0.3774 & 0.3492 & 0.3332          & 0.3268  & 0.3246  & 0.3224          & \textbf{0.3201} \\
		& Oct & 0.9891      & 0.5518 & 0.5032 & 0.5154 & 0.4693          & 0.5786 & 0.4141          & 0.4647 & 0.4652 & 0.4763 & \textbf{0.4345} & 0.4460  & 0.4449  & 0.4383          & 0.4354          \\
		TAS & Jan     & 1.1101      & 0.8751 & 1.0609 & 0.9565 & \textbf{0.8581} & 0.9171 & 0.8967          & 0.8819 & 0.8769 & 0.8633 & 0.8627          & 0.8594  & 0.8590  & 0.8601          & 0.8587          \\
		& Apr   & 1.0463      & 0.6983 & 0.6081 & 0.9746 & 0.6298          & 0.6694 & 0.5870          & 0.6926 & 0.6143 & 0.5968 & 0.6045          & 0.5803  & 0.5793  & 0.5756          & \textbf{0.5745} \\
		& Jul    & 1.1317      & 0.6349 & 0.5599 & 0.5926 & 0.5358          & 0.5890 & 0.5169          & 0.6721 & 0.5384 & 0.5500 & 0.5307          & 0.5078  & 0.5061  & 0.4932          & \textbf{0.4905} \\
		& Oct & 1.0218      & 0.6730 & 0.6162 & 0.6354 & 0.6145          & 0.6872 & 0.6295          & 0.6598 & 0.6269 & 0.6252 & 0.6210          & 0.6115  & 0.6106  & 0.6088          & \textbf{0.6083} \\
		\hline
	\end{tabular}}
\end{table*}

\begin{table*}[htbp]
		\centering
	\caption{Comparative results in terms of MAPE.}
	\label{tab:Comparative results in terms of MAPE.}
	\resizebox{\textwidth}{!}{
	\begin{tabular}{lllllllllllllllll}
		\hline
		Location & Month   & Persistence \cite{makridakis2008forecasting}& ARIMA \cite{contreras2003arima}  & SVR \cite{chen2004load}     & MLP \cite{kandil2006efficient}     & LSTM \cite{kong2017short}            & TCN \cite{baiempirical}   & EWTFCMSVR \cite{gao2020robust}        & WHFCM \cite{yang2018time}           & LapESN \cite{han2017laplacian}  & RVFL \cite{ren2016random}    & EWTRVFL \cite{gao2021walk}         & Med-edRVFL & Mea-edRVFL          & EWTMed-edRVFL       & EWTMea-edRVFL       \\
		\hline
		SA  & Jan     & 0.03832     & 0.02579 & 0.02579 & 0.02600 & 0.02413          & 0.02478 & 0.03112          & 0.02190          & 0.02313 & 0.02414 & 0.02143          & 0.02176 & 0.02178          & \textbf{0.02093} & 0.02101          \\
		& Apr   & 0.04442     & 0.03280 & 0.03104 & 0.03127 & 0.03330          & 0.03411 & 0.07170          & 0.03389          & 0.03246 & 0.03296 & 0.03098          & 0.03080 & 0.03065          & 0.03048          & \textbf{0.03034} \\
		& Jul    & 0.05192     & 0.02697 & 0.02229 & 0.02676 & 0.02013          & 0.02664 & 0.02053          & \textbf{0.02184} & 0.02584 & 0.02505 & 0.02228          & 0.02294 & 0.02270          & 0.02185          & 0.02185          \\
		& Oct & 0.04723     & 0.03363 & 0.02968 & 0.03962 & \textbf{0.02909} & 0.03218 & 0.04202          & 0.03016          & 0.03010 & 0.03037 & 0.03052          & 0.02932 & 0.02927          & 0.03116          & 0.03100          \\
		QLD & Jan     & 0.01639     & 0.00747 & 0.01072 & 0.00951 & 0.00688          & 0.00712 & 0.00707          & 0.00617          & 0.00628 & 0.00606 & 0.00589          & 0.00597 & 0.00596          & 0.00580          & \textbf{0.00577} \\
		& Apr   & 0.01848     & 0.00949 & 0.00714 & 0.00797 & 0.00600          & 0.01015 & 0.02166          & 0.00793          & 0.00811 & 0.00761 & 0.00725          & 0.00715 & 0.00709          & 0.00691          & \textbf{0.00683} \\
		& Jul    & 0.03002     & 0.01125 & 0.00899 & 0.01001 & 0.00818          & 0.00911 & 0.00842          & 0.00952          & 0.00877 & 0.00853 & \textbf{0.00789} & 0.00828 & 0.00826          & 0.00806          & 0.00800          \\
		& Oct & 0.01990     & 0.01176 & 0.01393 & 0.01271 & 0.01159          & 0.01161 & 0.02650          & 0.01072          & 0.01072 & 0.01101 & 0.01087          & 0.01060 & 0.01057          & 0.01057          & 0.01055          \\
		NSW & Jan     & 0.02287     & 0.00869 & 0.01373 & 0.01555 & 0.00865          & 0.00879 & 0.00859          & 0.00837          & 0.00854 & 0.00837 & 0.00833          & 0.00825 & \textbf{0.00824} & 0.00834          & 0.00833          \\
		& Apr   & 0.01901     & 0.01117 & 0.01001 & 0.00843 & 0.00984          & 0.01138 & 0.03066          & 0.00810          & 0.00914 & 0.00846 & 0.00729          & 0.00823 & 0.00817          & 0.00774          & \textbf{0.00765} \\
		& Jul    & 0.02753     & 0.01148 & 0.01074 & 0.01003 & 0.00914          & 0.00854 & \textbf{0.00765} & 0.00841          & 0.00917 & 0.00915 & 0.01012          & 0.00819 & 0.00842          & 0.00800          & 0.00797          \\
		& Oct & 0.02052     & 0.01015 & 0.01278 & 0.00988 & 0.01042          & 0.01277 & 0.01052          & 0.00887          & 0.00882 & 0.00891 & 0.00855          & 0.00843 & 0.00841          & 0.00813          & \textbf{0.00811} \\
		VIC & Jan     & 0.02269     & 0.01423 & 0.03252 & 0.01552 & 0.01743          & 0.01748 & \textbf{0.01102} & 0.01345          & 0.01293 & 0.01271 & 0.01115          & 0.01249 & 0.01252          & 0.01203          & 0.01199          \\
		& Apr   & 0.02973     & 0.01592 & 0.01832 & 0.01669 & 0.01457          & 0.01619 & 0.02435          & 0.01349          & 0.01348 & 0.01437 & 0.01442          & 0.01304 & 0.01305          & 0.01287          & \textbf{0.01283} \\
		& Jul    & 0.03014     & 0.01395 & 0.01068 & 0.01062 & 0.00950          & 0.01192 & 0.03815          & 0.01070          & 0.01097 & 0.01014 & 0.00975          & 0.00948 & 0.00942          & 0.00938          & \textbf{0.00931} \\
		& Oct & 0.02657     & 0.01496 & 0.01364 & 0.01411 & 0.01282          & 0.01570 & \textbf{0.01144} & 0.01267          & 0.01274 & 0.01303 & 0.01198          & 0.01220 & 0.01217          & 0.01201          & 0.01193          \\
		TAS & Jan     & 0.01633     & 0.01292 & 0.01551 & 0.01403 & 0.01267          & 0.01345 & 0.01326          & 0.01299          & 0.01294 & 0.01272 & 0.01272          & 0.01266 & \textbf{0.01265} & 0.01267          & \textbf{0.01265} \\
		& Apr   & 0.02101     & 0.01420 & 0.01247 & 0.02014 & 0.01292          & 0.01377 & 0.01205          & 0.01407          & 0.01251 & 0.01222 & 0.01243          & 0.01186 & 0.01185          & 0.01182          & \textbf{0.01179} \\
		& Jul    & 0.02673     & 0.01532 & 0.01356 & 0.01437 & 0.01299          & 0.01427 & 0.01260          & 0.01613          & 0.01304 & 0.01337 & 0.01292          & 0.01229 & 0.01226          & 0.01195          & \textbf{0.01189} \\
		& Oct & 0.02164     & 0.01442 & 0.01321 & 0.01363 & 0.01317          & 0.01464 & 0.01354          & 0.01409          & 0.01347 & 0.01344 & 0.01335          & 0.01314 & 0.01312          & 0.01308          & \textbf{0.01307} \\
		\hline
	\end{tabular}}
\end{table*}

\begin{table*}[htbp]
	\centering
	\caption{Average ranking of all models.}
	\label{tab:ranking.}
	\resizebox{\textwidth}{!}{
	\begin{tabular}{llllllllllllllll}
		\hline
		& Persistence \cite{makridakis2008forecasting} & ARIMA \cite{contreras2003arima} & SVR \cite{chen2004load}   & MLP \cite{kandil2006efficient}  & LSTM \cite{kong2017short} & TCN \cite{baiempirical}   & EWTFCMSVR \cite{gao2020robust} & WHFCM \cite{yang2018time} & LapESN \cite{han2017laplacian} & RVFL \cite{ren2016random} & EWTRVFL \cite{gao2021walk} & Med-edRVFL & Mea-edRVFLL & EWTMed-edRVFL & EWTMea-edRVFL \\
		\hline
		RMSE & 14.65       & 11.85 & 11.3  & 10.65 & 7.45 & 11.45 & 9.55      & 7.95  & 8.30    & 7.75 & 4.05    & 5.15    & 4.6     & 3.15       & 2.15       \\
		MASE & 14.7        & 11.85 & 10.6  & 11.1  & 7.4  & 11.7  & 9.45      & 8.25  & 8.25   & 7.85 & 4.75    & 4.7     & 3.95    & 3.25       & 2.20        \\
		MAPE & 14.70        & 11.80  & 10.35 & 11.40  & 7.20  & 11.7  & 9.55      & 8.1   & 8.25   & 7.90  & 5.35    & 4.5     & 3.9     & 3.15       & 2.15       \\
		\hline 
\end{tabular}}
\end{table*}
\begin{table*}[htbp]
		\centering
	\caption{Pairwise comparisons using Nemenyi post-hoc test based on RMSE.}
	\label{tab:Pairwise RMSE.}
	\resizebox{\textwidth}{!}{
	\begin{tabular}{llllllllllllllll}
		\hline
	            & Persistence \cite{makridakis2008forecasting}& ARIMA \cite{contreras2003arima}  & SVR \cite{chen2004load}    & MLP \cite{kandil2006efficient}    & LSTM \cite{kong2017short}   & TCN \cite{baiempirical}    & EWTFCMSVR \cite{gao2020robust} & WHFCM \cite{yang2018time} & LapESN \cite{han2017laplacian} & RVFL \cite{ren2016random}  & EWTRVFL \cite{gao2021walk}& Med-edRVFL & Mea-edRVFL & EWTMed-edRVFL & EWTMea-edRVFL \\
	            \hline
	Persistence \cite{makridakis2008forecasting}& -1.000      & 0.783  & 0.533  & 0.232  & 0.001  & 0.601  & 0.025     & 0.001  & 0.001  & 0.001  & 0.001   & 0.001   & 0.001   & 0.001      & 0.001      \\
	ARIMA \cite{contreras2003arima}      & 0.783       & -1.000 & 0.900  & 0.900  & 0.115  & 0.900  & 0.900     & 0.271  & 0.438  & 0.196  & 0.001   & 0.001   & 0.001   & 0.001      & 0.001      \\
	SVR \cite{chen2004load}        & 0.533       & 0.900  & -1.000 & 0.900  & 0.292  & 0.900  & 0.900     & 0.533  & 0.692  & 0.438  & 0.001   & 0.001   & 0.001   & 0.001      & 0.001      \\
	MLP \cite{kandil2006efficient}        & 0.232       & 0.900  & 0.900  & -1.000 & 0.601  & 0.900  & 0.900     & 0.829  & 0.900  & 0.738  & 0.001   & 0.009   & 0.002   & 0.001      & 0.001      \\
	LSTM \cite{kong2017short}       & 0.001       & 0.115  & 0.292  & 0.601  & -1.000 & 0.232  & 0.900     & 0.900  & 0.900  & 0.900  & 0.510   & 0.900   & 0.760   & 0.138      & 0.015      \\
	TCN \cite{baiempirical}        & 0.601       & 0.900  & 0.900  & 0.900  & 0.232  & -1.000 & 0.900     & 0.463  & 0.624  & 0.361  & 0.001   & 0.001   & 0.001   & 0.001      & 0.001      \\
	EWTFCMSVR \cite{gao2020robust}  & 0.025       & 0.900  & 0.900  & 0.900  & 0.900  & 0.900  & -1.000    & 0.900  & 0.900  & 0.900  & 0.009   & 0.115   & 0.035   & 0.001      & 0.001      \\
	WHFCM \cite{yang2018time}      & 0.001       & 0.271  & 0.533  & 0.829  & 0.900  & 0.463  & 0.900     & -1.000 & 0.900  & 0.900  & 0.271   & 0.783   & 0.533   & 0.050      & 0.004      \\
	LapESN \cite{han2017laplacian}      & 0.001       & 0.438  & 0.692  & 0.900  & 0.900  & 0.624  & 0.900     & 0.900  & -1.000 & 0.900  & 0.151   & 0.624   & 0.361   & 0.022      & 0.001      \\
	RVFL \cite{ren2016random}       & 0.001       & 0.196  & 0.438  & 0.738  & 0.900  & 0.361  & 0.900     & 0.900  & 0.900  & -1.000 & 0.361   & 0.874   & 0.624   & 0.077      & 0.007      \\
	EWTRVFL \cite{gao2021walk}    & 0.001       & 0.001  & 0.001  & 0.001  & 0.510  & 0.001  & 0.009     & 0.271  & 0.151  & 0.361  & -1.000  & 0.900   & 0.900   & 0.900      & 0.900      \\
	Med-edRVFL     & 0.001       & 0.001  & 0.001  & 0.009  & 0.900  & 0.001  & 0.115     & 0.783  & 0.624  & 0.874  & 0.900   & -1.000  & 0.900   & 0.900      & 0.692      \\
	Mea-edRVFL     & 0.001       & 0.001  & 0.001  & 0.002  & 0.760  & 0.001  & 0.035     & 0.533  & 0.361  & 0.624  & 0.900   & 0.900   & -1.000  & 0.900      & 0.900      \\
	EWTMed-edRVFL  & 0.001       & 0.001  & 0.001  & 0.001  & 0.138  & 0.001  & 0.001     & 0.050  & 0.022  & 0.077  & 0.900   & 0.900   & 0.900   & -1.000     & 0.900      \\
	EWTMea-edRVFL  & 0.001       & 0.001  & 0.001  & 0.001  & 0.015  & 0.001  & 0.001     & 0.004  & 0.001  & 0.007  & 0.900   & 0.692   & 0.900   & 0.900      & -1.000    \\
		\hline
	\end{tabular}}
\end{table*}
\begin{table*}[htbp]
	\centering
	\caption{Pairwise comparisons using Nemenyi post-hoc test based on MASE.}
	\label{tab:Pairwise MASE.}
	\resizebox{\textwidth}{!}{
	\begin{tabular}{llllllllllllllll}
		\hline
		& Persistence \cite{makridakis2008forecasting}& ARIMA \cite{contreras2003arima}  & SVR \cite{chen2004load}   & MLP \cite{kandil2006efficient}    & LSTM \cite{kong2017short}   & TCN \cite{baiempirical}    & EWTFCMSVR \cite{gao2020robust} & WHFCM \cite{yang2018time}  & LapESN \cite{han2017laplacian} & RVFL \cite{ren2016random}   & EWTRVFL \cite{gao2021walk} & Med-edRVFL & Mea-edRVFL & EWTMed-edRVFL & EWTMea-edRVFL \\
		\hline
		Persistence \cite{makridakis2008forecasting}& -1.000      & 0.760  & 0.196  & 0.412  & 0.001  & 0.692  & 0.017     & 0.001  & 0.001  & 0.001  & 0.001   & 0.001   & 0.001   & 0.001      & 0.001      \\
		ARIMA \cite{contreras2003arima}      & 0.760       & -1.000 & 0.900  & 0.900  & 0.104  & 0.900  & 0.900     & 0.412  & 0.412  & 0.232  & 0.001   & 0.001   & 0.001   & 0.001      & 0.001      \\
		SVR \cite{chen2004load}        & 0.196       & 0.900  & -1.000 & 0.900  & 0.601  & 0.900  & 0.900     & 0.900  & 0.900  & 0.806  & 0.003   & 0.003   & 0.001   & 0.001      & 0.001      \\
		MLP \cite{kandil2006efficient}        & 0.412       & 0.900  & 0.900  & -1.000 & 0.361  & 0.900  & 0.900     & 0.760  & 0.760  & 0.578  & 0.001   & 0.001   & 0.001   & 0.001      & 0.001      \\
		LSTM \cite{kong2017short}       & 0.001       & 0.104  & 0.601  & 0.361  & -1.000 & 0.138  & 0.900     & 0.900  & 0.900  & 0.900  & 0.851   & 0.829   & 0.487   & 0.180      & 0.019      \\
		TCN \cite{baiempirical}        & 0.692       & 0.900  & 0.900  & 0.900  & 0.138  & -1.000 & 0.900     & 0.487  & 0.487  & 0.292  & 0.001   & 0.001   & 0.001   & 0.001      & 0.001      \\
		EWTFCMSVR \cite{gao2020robust}   & 0.017       & 0.900  & 0.900  & 0.900  & 0.900  & 0.900  & -1.000    & 0.900  & 0.900  & 0.900  & 0.062   & 0.056   & 0.009   & 0.001      & 0.001      \\
		WHFCM \cite{yang2018time}      & 0.001       & 0.412  & 0.900  & 0.760  & 0.900  & 0.487  & 0.900     & -1.000 & 0.900  & 0.900  & 0.463   & 0.438   & 0.138   & 0.031      & 0.002      \\
		LapESN \cite{han2017laplacian}      & 0.001       & 0.412  & 0.900  & 0.760  & 0.900  & 0.487  & 0.900     & 0.900  & -1.000 & 0.900  & 0.463   & 0.438   & 0.138   & 0.031      & 0.002      \\
		RVFL \cite{ren2016random}       & 0.001       & 0.232  & 0.806  & 0.578  & 0.900  & 0.292  & 0.900     & 0.900  & 0.900  & -1.000 & 0.647   & 0.624   & 0.271   & 0.077      & 0.006      \\
		EWTRVFL \cite{gao2021walk}     & 0.001       & 0.001  & 0.003  & 0.001  & 0.851  & 0.001  & 0.062     & 0.463  & 0.463  & 0.647  & -1.000  & 0.900   & 0.900   & 0.900      & 0.897      \\
		Med-edRVFL     & 0.001       & 0.001  & 0.003  & 0.001  & 0.829  & 0.001  & 0.056     & 0.438  & 0.438  & 0.624  & 0.900   & -1.000  & 0.900   & 0.900      & 0.900      \\
		Mea-edRVFL     & 0.001       & 0.001  & 0.001  & 0.001  & 0.487  & 0.001  & 0.009     & 0.138  & 0.138  & 0.271  & 0.900   & 0.900   & -1.000  & 0.900      & 0.900      \\
		EWTMed-edRVFL  & 0.001       & 0.001  & 0.001  & 0.001  & 0.180  & 0.001  & 0.001     & 0.031  & 0.031  & 0.077  & 0.900   & 0.900   & 0.900   & -1.000     & 0.900      \\
		EWTMea-edRVFL  & 0.001       & 0.001  & 0.001  & 0.001  & 0.019  & 0.001  & 0.001     & 0.002  & 0.002  & 0.006  & 0.897   & 0.900   & 0.900   & 0.900      & -1.000     \\
		\hline
	\end{tabular}}
\end{table*}
\begin{table*}[]
		\centering
	\caption{Pairwise comparisons using Nemenyi post-hoc test based on MAPE.}
	\label{tab:Pairwise MAPE.}
	\resizebox{\textwidth}{!}{
	\begin{tabular}{llllllllllllllll}
		\hline
		& Persistence \cite{makridakis2008forecasting} & ARIMA \cite{contreras2003arima}  & SVR \cite{chen2004load}    & MLP \cite{kandil2006efficient}    & LSTM \cite{kong2017short}  & TCN \cite{baiempirical}    & EWTFCMSVR \cite{gao2020robust} & WHFCM \cite{yang2018time}  & LapESN \cite{han2017laplacian} & RVFL \cite{ren2016random}   & EWTRVFL \cite{gao2021walk} & Med-edRVFL & Mea-edRVFL & EWTMed-edRVFL & EWTMea-edRVFL \\
		\hline
		Persistence \cite{makridakis2008forecasting} & -1.000      & 0.738  & 0.125  & 0.556  & 0.001  & 0.692  & 0.022     & 0.001  & 0.001  & 0.001  & 0.001   & 0.001   & 0.001   & 0.001      & 0.001      \\
		ARIMA \cite{contreras2003arima}      & 0.738       & -1.000 & 0.900  & 0.900  & 0.077  & 0.900  & 0.900     & 0.361  & 0.438  & 0.271  & 0.001   & 0.001   & 0.001   & 0.001      & 0.001      \\
		SVR \cite{chen2004load}        & 0.125       & 0.900  & -1.000 & 0.900  & 0.624  & 0.900  & 0.900     & 0.900  & 0.900  & 0.900  & 0.031   & 0.003   & 0.001   & 0.001      & 0.001      \\
		MLP \cite{kandil2006efficient}        & 0.556       & 0.900  & 0.900  & -1.000 & 0.166  & 0.900  & 0.900     & 0.556  & 0.624  & 0.463  & 0.002   & 0.001   & 0.001   & 0.001      & 0.001      \\
		LSTM \cite{kong2017short}       & 0.001       & 0.077  & 0.624  & 0.166  & -1.000 & 0.094  & 0.900     & 0.900  & 0.900  & 0.900  & 0.900   & 0.829   & 0.556   & 0.214      & 0.028      \\
		TCN \cite{baiempirical}        & 0.692       & 0.900  & 0.900  & 0.900  & 0.094  & -1.000 & 0.900     & 0.412  & 0.487  & 0.313  & 0.001   & 0.001   & 0.001   & 0.001      & 0.001      \\
		EWTFCMSVR \cite{gao2020robust}  & 0.022       & 0.900  & 0.900  & 0.900  & 0.900  & 0.900  & -1.000    & 0.900  & 0.900  & 0.900  & 0.166   & 0.028   & 0.006   & 0.001      & 0.001      \\
		WHFCM \cite{yang2018time}      & 0.001       & 0.361  & 0.900  & 0.556  & 0.900  & 0.412  & 0.900     & -1.000 & 0.900  & 0.900  & 0.806   & 0.412   & 0.166   & 0.035      & 0.002      \\
		LapESN  \cite{han2017laplacian}     & 0.001       & 0.438  & 0.900  & 0.624  & 0.900  & 0.487  & 0.900     & 0.900  & -1.000 & 0.900  & 0.738   & 0.336   & 0.125   & 0.025      & 0.002      \\
		RVFL \cite{ren2016random}       & 0.001       & 0.271  & 0.900  & 0.463  & 0.900  & 0.313  & 0.900     & 0.900  & 0.900  & -1.000 & 0.897   & 0.510   & 0.232   & 0.056      & 0.004      \\
		EWTRVFL \cite{gao2021walk}    & 0.001       & 0.001  & 0.031  & 0.002  & 0.900  & 0.001  & 0.166     & 0.806  & 0.738  & 0.897  & -1.000  & 0.900   & 0.900   & 0.900      & 0.601      \\
		Med-edRVFL     & 0.001       & 0.001  & 0.003  & 0.001  & 0.829  & 0.001  & 0.028     & 0.412  & 0.336  & 0.510  & 0.900   & -1.000  & 0.900   & 0.900      & 0.900      \\
		Mea-edRVFL     & 0.001       & 0.001  & 0.001  & 0.001  & 0.556  & 0.001  & 0.006     & 0.166  & 0.125  & 0.232  & 0.900   & 0.900   & -1.000  & 0.900      & 0.900      \\
		EWTMed-edRVFL  & 0.001       & 0.001  & 0.001  & 0.001  & 0.214  & 0.001  & 0.001     & 0.035  & 0.025  & 0.056  & 0.900   & 0.900   & 0.900   & -1.000     & 0.900      \\
		EWTMea-edRVFL  & 0.001       & 0.001  & 0.001  & 0.001  & 0.028  & 0.001  & 0.001     & 0.002  & 0.002  & 0.004  & 0.601   & 0.900   & 0.900   & 0.900      & -1.000    \\
		\hline
	\end{tabular}}
\end{table*}
Table \ref{tab:time complexity.} records the simulation time for optimization and training time. It is worth noting that the optimization time is the time of the cross-validation using grid-search. The training time represents the time that the model is trained using the hyper-parameters selected by the cross-validation. The time for RVFL-related models is the summation of twenty runs. Several phenomenons are concluded according to Table \ref{tab:time complexity.}. The most time-consuming model is the LSTM because of its recurrent structure which processes the data in a sequential order. The hybrid RVFL model with EWT is more time-consuming than RVFL-related models. For example, the EWT-RVFL and EWT-edRVFL are more time-consuming than RVFL and edRVFL, respectively. Therefore, the main computation is in the walk-forward EWT decomposition block because it happens at each step. 
\begin{table}[]
	\centering
	\caption{Average time complexity (per second) for all the models.}
	\label{tab:time complexity.}
	\resizebox{.5\textwidth}{!}{
	\begin{tabular}{lll}
		\hline
		& Optimization time & Training time \\
		\hline
		ARIMA \cite{contreras2003arima}     & 42.595          & 3.692         \\
		SVR \cite{chen2004load}       & 4.058           & 0.109         \\
		MLP  \cite{kandil2006efficient}      & 65.260          & 4.386         \\
		LSTM \cite{kong2017short}      & 1561.631        & 150.642       \\
		TCN  \cite{baiempirical}      & 171.563         & 50.919        \\
		EWTFCMSVR \cite{gao2020robust} & 40.528          & 26.531        \\
		WHFCM \cite{yang2018time}     & 21.755          & 0.130         \\
		LapESN \cite{han2017laplacian}    & 29.182          & 6.078         \\
		RVFL    \cite{ren2016random}   & 1.689           & 0.140         \\
		EWTRVFL \cite{gao2021walk}   & 42.518          & 2.060         \\
		edRVFL     & 31.859          & 7.307         \\
		EWT-edRVFL & 75.620          & 14.067        \\
		\hline
	\end{tabular}}
\end{table}
\section{Conclusion}
This paper proposes a novel ensemble deep RVFL network combined with walk-forward decomposition for short-term load forecasting. The enhancement layers' weights are randomly initialized and kept fixed as in the shallow RVFL network. Only the output weights of each layer are computed in a closed form. Since the enhancement features are unsupervised and randomly initialized, the walk-forward EWT is implemented to augment the feature extraction. The walk-forward EWT is different from most literature, where the whole time series is decomposed at one time. Therefore, there is no data leakage problem during the decomposition process. Finally, the mean and median of all forecasts are used as the final output. The experiments on twenty electricity loads demonstrate the superiority and efficiency of the proposed model. Moreover, the proposed model does not suffer from a colossal computation burden compared with other deep learning models which are fully trained.

There are several reasons for the superiority of the proposed model:
\begin{enumerate}[1.]
\item The edRVFL's structure benefits from ensemble learning. The edRVFL treats each enhancement layer as a single forecaster. Therefore, the ensemble multiple forecasters reduce the uncertainty of a single forecaster.
\item The clean raw data are fed into all enhancement layers to calibrate the random features' generation.
\item The output layer learns both the linear patterns from the direct link and nonlinear patterns from the enhancement features.
\item The walk-forward EWT is used as a feature engineering block to boost the accuracy further.
\end{enumerate}

Although our model shows its superiority in these twenty datasets, there are still some limitations. For the walk-forward EWT process, whether to discard the highest frequency is an open problem. It is challenging to determine how valuable information is in the highest frequency component. Moreover, other learning techniques can be considered to further boost the performance, like incremental learning and semi-unsupervised learning.
\section*{Acknowledgment}

The authors thank the anonymous reviewers for providing
valuable comments to improve this paper.

\ifCLASSOPTIONcaptionsoff
  \newpage
\fi




\bibliographystyle{IEEEtran}
\bibliography{ref}
\end{document}